\documentclass{article}
\usepackage{ijcai25}

\usepackage{times}
\usepackage{soul}
\usepackage{url}
\usepackage[hidelinks]{hyperref}
\usepackage[utf8]{inputenc}
\usepackage[small]{caption}
\usepackage{graphicx}
\usepackage{amsmath}
\usepackage{amsthm}
\usepackage{amssymb}
\usepackage{booktabs}
\usepackage{algorithm}
\usepackage{algorithmic}
\usepackage[switch]{lineno}
\usepackage{multirow}
\usepackage{pifont}
\usepackage[table]{xcolor}

\newcommand{\cmark}{\ding{51}}
\newcommand{\xmark}{\ding{55}}
\newcommand{\mat}[1]{{\bf #1}}   

\title{Harnessing Vision Models for Time Series Analysis: A Survey}

\author{Jingchao Ni\textsuperscript{1}, Ziming Zhao\textsuperscript{1}, ChengAo Shen\textsuperscript{1}, Hanghang Tong\textsuperscript{2}, Dongjin Song\textsuperscript{3},\\Wei Cheng\textsuperscript{4}, Dongsheng Luo\textsuperscript{5}, Haifeng Chen\textsuperscript{4}
\affiliations
\textsuperscript{1}University of Houston, Houston, TX, USA\\
\textsuperscript{2}University of Illinois at Urbana-Champaign, Urbana, IL, USA\\
\textsuperscript{3}University of Connecticut, Storrs, CT, USA\\
\textsuperscript{4}NEC Laboratories America, Princeton, NJ, USA\\
\textsuperscript{5}Florida International University, Miami, FL, USA
\emails
\{jni7, zzhao35, cshen9\}@uh.edu, htong@illinois.edu,\\dongjin.song@uconn.edu, \{weicheng, haifeng\}@nec-labs.com, dluo@fiu.edu
}

\begin{document}

\maketitle

\begin{abstract}
Time series analysis has evolved from traditional autoregressive models to deep learning, Transformers, and Large Language Models (LLMs). While vision models have also been explored along the way, their contributions are less recognized due to the predominance of sequence modeling. However, challenges such as the mismatch between continuous time series and LLMs’ discrete token space, and the difficulty in capturing multivariate correlations, have led to growing interest in Large Vision Models (LVMs) and Vision-Language Models (VLMs). This survey highlights the advantages of vision models over LLMs in time series analysis, offering a comprehensive dual-view taxonomy that answers key research questions like how to encode time series as images and how to model imaged time series. Additionally, we address pre- and post-processing challenges in this framework and outline future directions for advancing the field.
\end{abstract}

\section{Introduction}\label{sec.introduction}

\begin{figure}[!t]
\centering
\includegraphics[width=0.77\columnwidth]{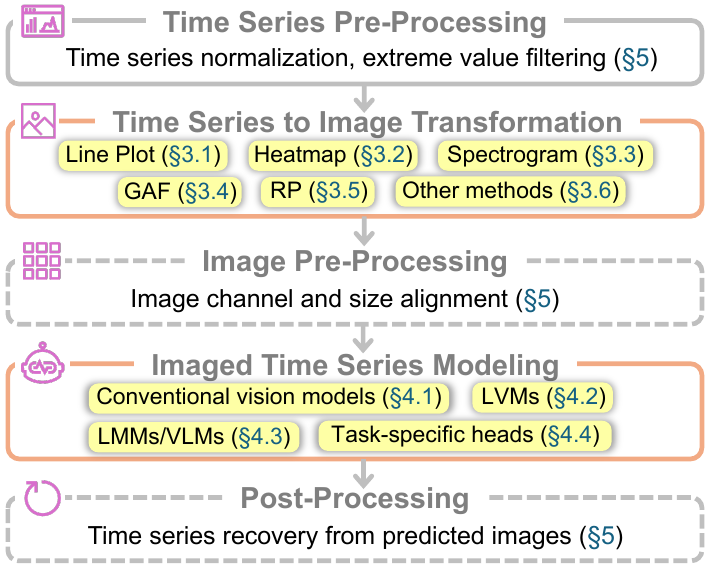}
\caption{The general process of leveraging vision models for time series analysis. The red boxes are two views of taxonomy used in this survey. The dashed boxes denote optional, task-dependent steps.}\label{fig.structure}
\end{figure}

Vision models have long been applied to time series analysis. Early successes with 1D CNNs \cite{bai2018empirical}, such as WaveNet \cite{van2016wavenet}, led to their widespread use across time series tasks \cite{koprinska2018convolutional,zhang2020tapnet}. Recently, advances in language-based sequence modeling have shifted attention to Transformers \cite{wen2023transformers} and Large Language Models (LLMs) \cite{zhang2024large}. Meanwhile, the demands for universal modeling has driven the emergence of time series foundation models such as TimesFM \cite{das2024decoder}, Chronos \cite{ansari2024chronos}, and Time-MoE \cite{shi2024time}.

As Large Vision Models (LVMs), such as ViT \cite{dosovitskiy2021image}, 
BEiT \cite{bao2022beit} and MAE \cite{he2022masked}, become achieving a similar success as LLMs (but in vision domain), a great deal of emergent efforts has been invested to explore the potential of LVMs in time series modeling \cite{chen2024visionts}. This is inspired by the plenty of ways for visualizing time series as images such as line plots of univariate time series (UTS) and heatmaps of multivariate time series (MTS). Such images provide a more straightforward view of time series than the counterpart textual representations to humans and, presumably, AI bots.

Taking a closer inspection reveals more advantages favoring LVMs over LLMs: (1) There is an inherent relationship between images and time series -- each row/column in an image (per channel) is a sequence of {\em continuous} pixel values. By pre-training on massive images, LVMs may have learned important sequential patterns such as trends, periods, and spikes \cite{chen2024visionts}. In contrast, LLMs are pre-trained on {\em discrete} tokens, thus are less aligned with continuous time series. In fact, LLMs' effectiveness on time series modeling is in question \cite{tan2024language}; (2) Instead of using channel-independence assumption \cite{nie2023time} to individually model each variate in an MTS, some imaging methods ($\S$\ref{sec.modelmts}) can naturally represent MTS, enabling explicit correlation encoding; (3) When prompting LLMs, existing methods often struggle with properly 
verbalizing a long sequence (or a matrix) of floating numbers in a UTS (or MTS), which may be limited by the context length or induce high API costs. In contrast, existing works find that using LVMs on imaged time series is more prompt-friendly and less API-costly \cite{daswani2024plots}; (4) Some imaging methods 
can encode long time series in a compact manner \cite{naiman2024utilizing}, thus have a great potential in modeling long-term dependency.

Also, the concurrent developments of LLMs and LVMs for time series 
pave the way for a confluence, 
{\em i.e.}, leveraging 
Large Multimodal Models (LMMs), 
such as LLaVA \cite{liu2023visual}, 
Gemini \cite{team2023gemini} 
and Claude-3 \cite{anthropic2024claude}, to consolidate the two complementary modalities, which may 
revolutionize the way ({\em e.g.}, visually, linguistically, {\em etc.}) that users interact with time series.

Despite the significance, 
a thorough review of 
relevant works is absent in the existing literature to the best of our knowledge. The survey \cite{zhang2024large} discusses a few vision models, but its focus is 
LLMs for time series. In light of this, in this survey, we comprehensively investigate the 
traditional and the state-of-the-art (SOTA) methods. 
Fig. \ref{fig.structure} identifies the general process of applying vision models for time series analysis, which also serves as the structure of this survey. 
Our taxonomy has a dual view: (1) in Time Series to Image Transformation ($\S$\ref{sec.tsimage}), we review 5 primary {\em imaging methods} including Line Plot, Heatmap, Spectrogram, Gramian Angular Field (GAF), Recurrence Plot (RP), and some other 
methods; (2) in Imaged Time Series Modeling ($\S$\ref{sec.model}), we discuss conventional vision models, LVMs and the initial efforts in LMMs. To highlight the taxonomy, we defer the discussion on the 
desiderata of pre- and post-processing to the end of this survey ($\S$\ref{sec.processing}). For comparison, we provide Table \ref{tab.taxonomy} to 
summarize the existing methods. Finally, we discuss future directions 
in this 
promising field ($\S$\ref{sec.future}). A Github repository\footnote{\url{https://github.com/D2I-Group/awesome-vision-time-series}} is also maintained to provide up-to-date resources including our code of the imaging methods in $\S$\ref{sec.tsimage}. We hope this survey could be an orthogonal complement to the existing surveys on Transformer \cite{wen2023transformers}, LLMs \cite{zhang2024large,jiang2024empowering} and foundation models \cite{liang2024foundation} for time series, and 
provide a complete view on the process of using vision models for time series analysis, so as to be an insightful guidebook to the developers in this area.

\section{Preliminaries and Taxonomy}

In this paper, a UTS is represented by $\mat{x}=[x_{1}, ..., x_{T}]\in\mathbb{R}^{1\times T}$ where $T$ is the length of the UTS, $x_{t}$ ($1\le t\le T)$ is the value at time step $t$. 
Suppose there are $d$ variates (or features), let $\mat{x}_{i}\in\mathbb{R}^{1\times T}$ ($1\le i\le d$) be a UTS of the $i$-th variate, an MTS can be represented by $\mat{X}=[\mat{x}_{1}^{\top}, ..., \mat{x}_{d}^{\top}]^{\top}\in\mathbb{R}^{d\times T}$.

As illustrated in Fig. \ref{fig.structure}, this survey focuses on methods that transform time series to images, namely {\em imaged time series}, and then apply vision models 
on the imaged time series for tackling time series tasks, such as classification, forecasting and anomaly detection. It is noteworthy that methods on videos or sequential images ({\em a.k.a.} image time series \cite{tarasiou2023vits}) do not belong to this category because they don't transform time series to images. Similarly, methods for spaciotemporal traffic data are out of our scope if the methods focus on streams of images ({\em e.g.}, traffic flows in a stream of grid maps \cite{zhang2017deep}), but methods on imaging time-space matrices \cite{ma2017learning} that resemble MTS are included. 
For vision models on audios, this survey 
only discusses 
some representative works in $\S$\ref{sec.spectrogram} due to space limit. 
The focus of the survey will remain on general time series.

\subsection{Taxonomy}

We propose a taxonomy from 
the two views of {\em Time Series to Image Transformation} ($\S$\ref{sec.tsimage}) and {\em Imaged Time Series Modeling} ($\S$\ref{sec.model}) as illustrated in Fig. \ref{fig.structure}. For the former, 
we 
discuss 5 primary methods for 
imaging UTS or MTS, 
and remark on their pros and cons. For the latter, 
we classify the existing methods by conventional vision models, LVMs and LMMs. We discuss their strategies on pre-training, fine-tuning, prompting, and the deigns of task-specific heads. We also discuss the challenges and solutions in pre-/post-processing in $\S$\ref{sec.processing}. Table \ref{tab.taxonomy} presents a summary. In the following two sections, we will delve into the existing methods from the two views.


\begin{figure*}[!t]
\centering
\includegraphics[width=1.0\textwidth]{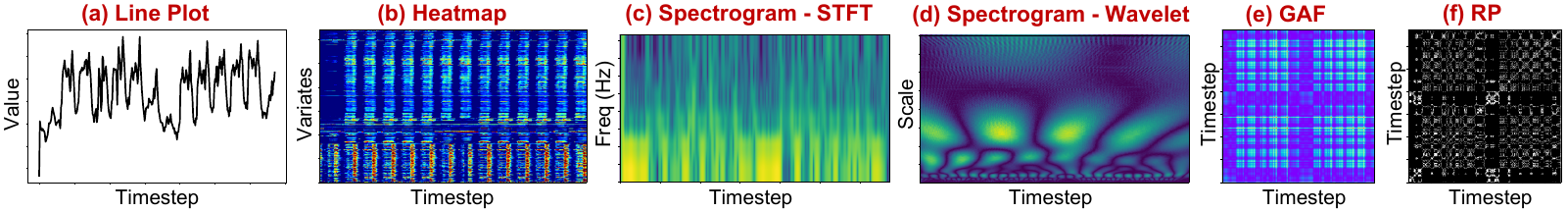}
\caption{An illustration of different methods for imaging time series with a sample (length=336) from the \textit{Electricity} benchmark dataset \protect\cite{nie2023time}. (a)(c)(d)(e)(f) 
visualize the same variate. (b) visualizes all 321 variates. Filterbank is omitted due to its 
similarity to STFT.}\label{fig.tsimage}
\end{figure*}

\begin{table*}[t]
\centering
\scriptsize
\setlength{\tabcolsep}{2.7pt}{
\begin{tabular}{llcccccccccl}
\toprule[1pt]
\multirow{2}{*}{Method} & \multirow{2}{*}{TS-Type} & \multirow{2}{*}{Imaging} & \multicolumn{5}{c}{Imaged Time Series Modeling} & \multirow{2}{*}{TS-Recover} & \multirow{2}{*}{Task} & \multirow{2}{*}{Domain} & \multirow{2}{*}{Code}\\ \cmidrule{4-8}
 & & & Multimodal & Model & Pre-trained & Fine-tune & Prompt & & & & \\ \midrule
\cite{silva2013time} & UTS & RP & \xmark & \texttt{K-NN} & \xmark & \xmark & \xmark & \xmark & Classification & General & \xmark\\
\cite{wang2015encoding} & UTS & GAF & \xmark & \texttt{CNN} & \xmark & \cmark$^{\flat}$ & \xmark & \cmark & Classification & General & \xmark\\
\cite{wang2015imaging} & UTS & GAF & \xmark & \texttt{CNN} & \xmark & \cmark$^{\flat}$ & \xmark & \cmark & Multiple & General & \xmark\\
\cite{ma2017learning} & MTS & Heatmap & \xmark & \texttt{CNN} & \xmark & \cmark$^{\flat}$ & \xmark & \cmark & Forecasting & Traffic & \xmark\\
\cite{hatami2018classification} & UTS & RP & \xmark & \texttt{CNN} & \xmark & \cmark$^{\flat}$ & \xmark & \xmark & Classification & General & \xmark\\
\cite{yazdanbakhsh2019multivariate} & MTS & Heatmap & \xmark & \texttt{CNN} & \xmark & \cmark$^{\flat}$ & \xmark & \xmark & Classification & General & \cmark\textsuperscript{\href{https://github.com/SonbolYb/multivariate_timeseries_dilated_conv}{[1]}}\\
MSCRED \cite{zhang2019deep} & MTS & Other ($\S$\ref{sec.othermethod}) & \xmark & \texttt{ConvLSTM} & \xmark & \cmark$^{\flat}$ & \xmark & \xmark & Anomaly & General & \cmark\textsuperscript{\href{https://github.com/7fantasysz/MSCRED}{[2]}}\\
\cite{li2020forecasting} & UTS & RP & \xmark & \texttt{CNN} & \cmark & \cmark & \xmark & \xmark & Forecasting & General & \cmark\textsuperscript{\href{https://github.com/lixixibj/forecasting-with-time-series-imaging}{[3]}}\\
\cite{cohen2020trading} & UTS & LinePlot & \xmark & \texttt{Ensemble} & \xmark & \cmark$^{\flat}$ & \xmark & \xmark & Classification & Finance & \xmark\\
\cite{barra2020deep} & UTS & GAF & \xmark & \texttt{CNN} & \xmark & \cmark$^{\flat}$ & \xmark & \xmark & Classification & Finance & \xmark\\
VisualAE \cite{sood2021visual} & UTS & LinePlot & \xmark & \texttt{CNN} & \xmark & \cmark$^{\flat}$ & \xmark & \cmark & Forecasting & Finance & \xmark\\
\cite{zeng2021deep} & MTS & Heatmap & \xmark & \texttt{CNN,LSTM} & \xmark & \cmark$^{\flat}$ & \xmark & \cmark & Forecasting & Finance & \xmark\\
AST \cite{gong2021ast} & UTS & Spectrogram & \xmark & \texttt{DeiT} & \cmark & \cmark & \xmark & \xmark & Classification & Audio & \cmark\textsuperscript{\href{https://github.com/YuanGongND/ast}{[4]}}\\
TTS-GAN \cite{li2022tts} & MTS & Heatmap & \xmark & \texttt{ViT} & \xmark & \cmark$^{\flat}$ & \xmark & \cmark & Ts-Generation & Health & \cmark\textsuperscript{\href{https://github.com/imics-lab/tts-gan}{[5]}}\\
SSAST \cite{gong2022ssast} & UTS & Spectrogram & \xmark & \texttt{ViT} & \cmark$^{\natural}$ & \cmark & \xmark & \xmark & Classification & Audio & \cmark\textsuperscript{\href{https://github.com/YuanGongND/ssast}{[6]}}\\
MAE-AST \cite{baade2022mae} & UTS & Spectrogram & \xmark & \texttt{MAE} & \cmark$^{\natural}$ & \cmark & \xmark & \xmark & Classification & Audio & \cmark\textsuperscript{\href{https://github.com/AlanBaade/MAE-AST-Public}{[7]}}\\
AST-SED \cite{li2023ast} & UTS & Spectrogram & \xmark & \texttt{SSAST,GRU} & \cmark & \cmark & \xmark & \xmark & EventDetection & Audio & \xmark\\
ForCNN \cite{semenoglou2023image} & UTS & LinePlot & \xmark & \texttt{CNN} & \xmark & \cmark$^{\flat}$ & \xmark & \xmark & Forecasting & General & \xmark\\
Vit-num-spec \cite{zeng2023pixels} & UTS & Spectrogram & \xmark & \texttt{ViT} & \xmark & \cmark$^{\flat}$ & \xmark & \xmark & Forecasting & Finance & \xmark\\
ViTST \cite{li2023time} & MTS & LinePlot & \xmark & \texttt{Swin} & \cmark & \cmark & \xmark & \xmark & Classification & General & \cmark\textsuperscript{\href{https://github.com/Leezekun/ViTST}{[8]}}\\
MV-DTSA \cite{yang2023your} & UTS\textsuperscript{*} & LinePlot & \xmark & \texttt{CNN} & \xmark & \cmark$^{\flat}$ & \xmark & \cmark & Forecasting & General & \cmark\textsuperscript{\href{https://github.com/IkeYang/machine-vision-assisted-deep-time-series-analysis-MV-DTSA-}{[9]}}\\
TimesNet \cite{wu2023timesnet} & MTS & Heatmap & \xmark & \texttt{CNN} & \xmark & \cmark$^{\flat}$ & \xmark & \cmark & Multiple & General & \cmark\textsuperscript{\href{https://github.com/thuml/TimesNet}{[10]}}\\
ITF-TAD \cite{namura2024training} & UTS & Spectrogram & \xmark & \texttt{CNN} & \cmark & \xmark & \xmark & \xmark & Anomaly & General & \xmark\\
\cite{kaewrakmuk2024multi} & UTS & GAF & \xmark & \texttt{CNN} & \cmark & \cmark & \xmark & \xmark & Classification & Sensing & \xmark\\
HCR-AdaAD \cite{lin2024hierarchical} & MTS & RP & \xmark & \texttt{CNN,GNN} & \xmark & \cmark$^{\flat}$ & \xmark & \xmark & Anomaly & General & \xmark\\
FIRTS \cite{costa2024fusion} & UTS & Other ($\S$\ref{sec.othermethod}) & \xmark & \texttt{CNN} & \xmark & \cmark$^{\flat}$ & \xmark & \xmark & Classification & General & \cmark\textsuperscript{\href{https://sites.google.com/view/firts-paper}{[11]}}\\
CAFO \cite{kim2024cafo} & MTS & RP & \xmark & \texttt{CNN,ViT} & \xmark & \cmark$^{\flat}$ & \xmark & \xmark & Explanation & General & \cmark\textsuperscript{\href{https://github.com/eai-lab/CAFO}{[12]}}\\
ViTime \cite{yang2024vitime} & UTS\textsuperscript{*} & LinePlot & \xmark & \texttt{ViT} & \cmark$^{\natural}$ & \cmark & \xmark & \cmark & Forecasting & General & \cmark\textsuperscript{\href{https://github.com/IkeYang/ViTime}{[13]}}\\
ImagenTime \cite{naiman2024utilizing} & MTS & Other ($\S$\ref{sec.othermethod}) & \xmark & 
\texttt{CNN} & \xmark & \cmark$^{\flat}$ & \xmark & \cmark & Ts-Generation & General & \cmark\textsuperscript{\href{https://github.com/azencot-group/ImagenTime}{[14]}}\\
TimEHR \cite{karami2024timehr} & MTS & Heatmap & \xmark & \texttt{CNN} & \xmark & \cmark$^{\flat}$ & \xmark & \cmark & Ts-Generation & Health & \cmark\textsuperscript{\href{https://github.com/esl-epfl/TimEHR}{[15]}}\\
VisionTS \cite{chen2024visionts} & UTS\textsuperscript{*} & Heatmap & \xmark & \texttt{MAE} & \cmark & \cmark & \xmark & \cmark & Forecasting & General & \cmark\textsuperscript{\href{https://github.com/Keytoyze/VisionTS}{[16]}}\\
TimeMixer++ \cite{wang2025timemixer} & MTS & Heatmap & \xmark & \texttt{CNN} & \xmark & \cmark$^{\flat}$ & \xmark & \cmark & Multiple & General & \cmark\textsuperscript{\href{https://anonymous.4open.science/r/TimeMixerPP}{[17]}}\\ \midrule
InsightMiner \cite{zhang2023insight} & UTS & LinePlot & \cmark & \texttt{LLaVA} & \cmark & \cmark & \cmark & \xmark & Txt-Generation & General & \xmark\\
\cite{wimmer2023leveraging} & MTS & LinePlot & \cmark & \texttt{CLIP,LSTM} & \cmark & \cmark & \xmark & \xmark & Classification & Finance & \xmark\\
\multirow{2}{*}{\cite{dixit2024vision}} & \multirow{2}{*}{UTS} & \multirow{2}{*}{Spectrogram} & \multirow{2}{*}{\cmark} & \texttt{GPT4o,Gemini} & \multirow{2}{*}{\cmark} & \multirow{2}{*}{\xmark} & \multirow{2}{*}{\cmark} & \multirow{2}{*}{\xmark} & \multirow{2}{*}{Classification} & \multirow{2}{*}{Audio} & \multirow{2}{*}{\xmark}\\
 & & & & \& \texttt{Claude3} & & & & & & & \\
\cite{daswani2024plots} & MTS & LinePlot & \cmark & \texttt{GPT4o,Gemini} & \cmark & \xmark & \cmark & \xmark & Multiple & General & \xmark\\
TAMA \cite{zhuang2024see} & UTS & LinePlot & \cmark & \texttt{GPT4o} & \cmark & \xmark & \cmark & \xmark & Anomaly & General & \xmark\\
\cite{prithyani2024feasibility} & MTS & LinePlot & \cmark & \texttt{LLaVA} & \cmark & \cmark & \cmark & \xmark & Classification & General & \cmark\textsuperscript{\href{https://github.com/vinayp17/VLM_TSC}{[18]}}\\
\bottomrule[1pt]
\end{tabular}}
\caption{Taxonomy of vision models on time series. The top part includes unimodal models. The bottom part includes multimodal models. {\bf TS-Type} denotes type of time series. {\bf TS-Recover} denotes 
recovering time series from predicted images ($\S$\ref{sec.processing}). \textsuperscript{*}: 
the method has been used to model the individual variates of an MTS. $^{\natural}$: a new pre-trained model was proposed in the work. $^{\flat}$: 
when pre-trained models were unused, ``Fine-tune'' refers to train a task-specific model from scratch. 
{\bf Model} column: \texttt{CNN} could be regular CNN, ResNet, VGG-Net, 
{\em etc.}}\label{tab.taxonomy}
\end{table*}

\begin{table*}[t]
\centering
\small
\setlength{\tabcolsep}{2.9pt}{
\begin{tabular}{l|l|l|l}\hline
\rowcolor{gray!20}
{\bf Method} & {\bf TS-Type} & {\bf Advantages} & {\bf Limitations}\\ \hline
Line Plot ($\S$\ref{sec.lineplot}) & UTS, MTS & matches human perception of time series & limited to MTS with a small number of variates\\ \hline
Heatmap ($\S$\ref{sec.heatmap}) & UTS, MTS & straightforward for both UTS and MTS & the order of variates may affect their correlation learning\\ \hline
Spectrogram ($\S$\ref{sec.spectrogram}) & UTS & encodes the time-frequency space & limited to UTS; needs a proper choice of window/wavelet\\ \hline
GAF ($\S$\ref{sec.gaf}) & UTS & encodes the temporal correlations in a UTS & limited to UTS; $O(T^{2})$ time and space complexity\\ \hline
RP ($\S$\ref{sec.rp}) & UTS & flexibility in image size by tuning $m$ and $\tau$ & limited to UTS; information loss after thresholding\\ \hline
\end{tabular}}
\caption{Summary of the five primary methods for transforming time series to images. {\bf TS-Type} denotes type of time series.}\label{tab.tsimage}
\end{table*}

\section{Time Series To Image Transformation}\label{sec.tsimage}

This section summarizes the methods for imaging time series ($\S$\ref{sec.lineplot}-$\S$\ref{sec.othermethod}) and their extensions to encode MTS ($\S$\ref{sec.modelmts}).

\subsection{Line Plot}\label{sec.lineplot}

Line Plot is a straightforward way for visualizing UTS for human analysis ({\em e.g.}, stocks, power consumption, {\em etc.}). As illustrated by Fig. \ref{fig.tsimage}(a), the simplest approach is to draw a 2D image with x-axis representing 
time steps and y-axis representing 
time-wise values, 
with a line connecting all values of the series over time. This image can be 
either three-channel ({\em i.e.}, RGB) or single-channel as the colors may not 
be informative 
\cite{cohen2020trading,sood2021visual,zhang2023insight}. ForCNN \cite{semenoglou2023image} even uses a single 8-bit integer to represent each pixel for black-white images. So far, there is no consensus on whether other graphical components, such as legend, grids and tick labels, could provide extra benefits in any task. For example, ViTST \cite{li2023time} finds these components are superfluous in a classification task, while TAMA \cite{zhuang2024see} finds grid-like auxiliary lines help enhance anomaly detection.

In addition to the regular Line Plot, MV-DTSA \cite{yang2023your} and ViTime \cite{yang2024vitime} divide an image into $h\times L$ grids, 
and 
define a function to map each time step of a UTS to a grid, producing a grid-like Line Plot. Also, we include methods that use Scatter Plot \cite{daswani2024plots,prithyani2024feasibility} in this category because 
a Scatter Plot resembles a Line Plot but doesn't connect 
data points with a line. By comparing them, \cite{prithyani2024feasibility} finds a Line Plot could induce better time series classification.

For MTS, we defer the discussion on Line Plot to $\S$\ref{sec.modelmts}.

\subsection{Heatmap}\label{sec.heatmap}

As shown in Fig. \ref{fig.tsimage}(b), a Heatmap visualizes the magnitudes of the values in matrix using color. It has been used to represent the matrix of MTS, {\em i.e.}, $\mat{X} \in \mathbb{R}^{d\times T}$, as a one-channel $d\times T$ image \cite{li2022tts,yazdanbakhsh2019multivariate}. TimEHR \cite{karami2024timehr} extends this to {\em irregular} MTS, where the intervals between time steps are uneven, by grouping the uneven time steps into $H$ uniform bins, creating a $d\times H$ Heatmap image. In \cite{zeng2021deep}, a different method is used for visualizing a 9-variate financial MTS. It reshapes the 9 variates at each time step to a $3\times 3$ Heatmap, and uses a sequence of the images to forecast future frames, achieving time series forecasting. In contrast, VisionTS \cite{chen2024visionts} uses Heatmap to visualize UTS. Similar to TimesNet \cite{wu2023timesnet} and TimeMixer++ \cite{wang2025timemixer}, it segments a length-$T$ UTS into $\lfloor T/P\rfloor$ length-$P$ subsequences, where $P$ denotes a period, stacks them into a $P\times \lfloor T/P\rfloor$ matrix, and replicates the channel 3 times to form a grayscale image for input to an LVM. For MTS, VisionTS adopts channel-independence strategy \cite{nie2023time} and individually models each variate.

\vspace{0.2cm}

\noindent{\bf Remark.} Heatmap can be used to visualize matrices of various forms. It is also used for matrices generated by the subsequent methods ({\em e.g.}, Spectrogram, GAF, RP) in this section. In this paper, the name Heatmap refers specifically to images that use color to visualize the (normalized) values in UTS $\mat{x}$ or MTS $\mat{X}$ without performing other transformations.

\subsection{Spectrogram}\label{sec.spectrogram}

A {\em spectrogram} is a visual representation of the spectrum of frequencies of a signal as it varies with time, which are extensively used for analyzing audio signals \cite{gong2021ast}. Since audio signals are a type of UTS, spectrogram can be considered as a method for imaging a UTS. As shown in Fig. \ref{fig.tsimage}(c), a common format is a 2D heatmap image with x-axis representing time steps and y-axis representing frequency, {\em a.k.a.} a time-frequency space. 
Each pixel in the image represents the (logarithmic) amplitude of a specific frequency at a specific time point. Typical methods for 
producing a spectrogram include {\bf Short-Time Fourier Transform (STFT)} \cite{griffin1984signal}, {\bf Wavelet Transform} \cite{daubechies1990wavelet}, and {\bf Filterbank} \cite{vetterli1992wavelets}.

\vspace{0.2cm}

\noindent{\bf STFT.} Discrete Fourier transform (DFT) can be used to describe the intensity $f(w)$ of each constituent frequency $w$ of a UTS signal $\mat{x}\in\mathbb{R}^{1\times T}$. However, $f(w)$ has no time dependency. It cannot provide dynamic information such as when a specific frequency appears in the UTS. STFT addresses this deficiency by sliding a window function $g(t)$ over the time steps in 
$\mat{x}$, and computing the DFT within each window by
\begin{equation}\label{eq.stft}
\small
\begin{aligned}
f(w,\tau) = \sum_{t=1}^{T}x_{t}g(t - \tau)e^{-iwt}
\end{aligned}
\end{equation}
where $w$ is frequency, $\tau$ is the position of the window, $f(w,\tau)$ describes the intensity of frequency $w$ at time step $\tau$.

By selecting a proper window function $g(\cdot)$ ({\em e.g.}, Gaussian/Hamming/Bartlett window), 
a 2D spectrogram ({\em e.g.}, Fig. \ref{fig.tsimage}(c)) can be drawn via a heatmap on the squared values $|f(w,\tau)|^{2}$, with $w$ as the y-axis, and $\tau$ as the x-axis. For example, \cite{dixit2024vision} uses STFT based spectrogram as an input to LMMs 
for time series classification.

\vspace{0.2cm}

\noindent{\bf Wavelet Transform.} 
Continuous Wavelet Transform (CWT) uses the inner product to measure the similarity between a signal function $x(t)$ and an analyzing function. 
The analyzing function is a {\em wavelet} $\psi(t)$, where the typical choices include Morse wavelet, Morlet wavelet, 
{\em etc.} 
CWT compares $x(t)$ to the shifted and scaled ({\em i.e.}, stretched or shrunk) versions of the wavelet, and output a CWT coefficient by
\begin{equation}\label{eq.cwt}
\small
\begin{aligned}
c(s,\tau) = \int_{-\infty}^{\infty}x(t)\frac{1}{s}\psi^{*}(\frac{t - \tau}{s})dt
\end{aligned}
\end{equation}
where $*$ denotes complex conjugate, $\tau$ is the time step to shift, and $s$ represents the scale. In practice, a discretized version of CWT in Eq.~\eqref{eq.cwt} is implemented for UTS $[x_{1}, ..., x_{T}]$.

It is noteworthy that the scale $s$ controls the frequency encoded in a wavelet -- a larger $s$ leads to a stretched wavelet with a lower frequency, and vice versa. As such, by varying $s$ and $\tau$, a 2D spectrogram ({\em e.g.}, Fig. \ref{fig.tsimage}(d)) can be drawn 
on $|c(s,\tau)|$, where $s$ is the y-axis and $\tau$ is the x-axis. Compared to STFT, which uses a fixed window size, Wavelet Transform allows variable wavelet sizes -- a larger size 
for more precise low frequency information. Thus, the methods in \cite{du2020image,namura2024training,zeng2023pixels} choose CWT (with Morlet wavelet) to generate the spectrogram.

\vspace{0.2cm}

\noindent{\bf Filterbank.} This method 
resembles STFT and is often used in processing audio signals. Given an audio signal, it firstly goes through a {\em pre-emphasis filter} to boost high frequencies, which helps improve the clarity of the signal. Then, STFT is applied on the signal. 
Finally, multiple ``triangle-shaped'' filters spaced on a Mel-scale are applied to the STFT power spectrum $|f(w, \tau)|^{2}$ to extract frequency bands. The outcome filterbank features $\hat{f}(w, \tau)$ can be used to yield a spectrogram with $w$ as the y-axis, and $\tau$ as the x-axis.

Filterbank was adopted in AST \cite{gong2021ast} with 
a 25ms Hamming window that shifts every 10ms for classifying audio signals using Vision Transformer (ViT). It then becomes widely used in the follow-up works such as SSAST \cite{gong2022ssast}, MAE-AST \cite{baade2022mae}, and AST-SED \cite{li2023ast}, as summarized in Table \ref{tab.taxonomy}.




\subsection{Gramian Angular Field (GAF)}\label{sec.gaf}

GAF was introduced for classifying UTS using CNNs 
by \cite{wang2015encoding}. It was then extended 
to an imputation task in \cite{wang2015imaging}. Similarly, \cite{barra2020deep} applied GAF for financial time series forecasting.

Given a UTS $\mat{x}\in\mathbb{R}^{1\times T}$, 
the first step 
is to rescale each $x_{t}$ to a value $\tilde{x}_{t}$ 
within $[0, 1]$ (or $[-1, 1]$). 
This range enables mapping $\tilde{x}_{t}$ to polar coordinates by $\phi_{t}=\text{arccos}(\tilde{x}_{i})$, with a radius $r=t/N$ encoding the time stamp, where $N$ is a constant factor to regularize the span of the polar coordinates. 
Two types of GAF, Gramian Sum Angular Field (GASF) and Gramian Difference Angular Field (GADF) are defined as
\begin{equation}\label{eq.gaf}
\small
\begin{aligned}
&\text{GASF:}~~\text{cos}(\phi_{t} + \phi_{t'})=x_{t}x_{t'} - \sqrt{1 - x_{t}^{2}}\sqrt{1 - x_{t'}^{2}}\\
&\text{GADF:}~~\text{sin}(\phi_{t} - \phi_{t'})=x_{t'}\sqrt{1 - x_{t}^{2}} - x_{t}\sqrt{1 - x_{t'}^{2}}
\end{aligned}
\end{equation}
which exploits the pairwise temporal correlations in the UTS. Thus, the outcome is a $T\times T$ matrix $\mat{G}$ with $\mat{G}_{t,t'}$ specified by either type in Eq.~\eqref{eq.gaf}. A GAF image is a heatmap on $\mat{G}$ with both axes representing time, as illustrated by Fig. \ref{fig.tsimage}(e).



\subsection{Recurrence Plot (RP)}\label{sec.rp}


RP \cite{eckmann1987recurrence} encodes a UTS into an image that captures its periodic patterns by using its reconstructed {\em phase space}. The phase space of 
$\mat{x}\in\mathbb{R}^{1\times T}$ can be reconstructed by {\em time delay embedding} -- a set of new vectors $\mat{v}_{1}$, ..., $\mat{v}_{l}$ with
\begin{equation}\label{eq.de}
\small
\begin{aligned}
\mat{v}_{t}=[x_{t}, x_{t+\tau}, x_{t+2\tau}, ..., x_{t+(m-1)\tau}]\in\mathbb{R}^{m\tau},~~~1\le t \le l
\end{aligned}
\end{equation}
where $\tau$ is the time delay, $m$ is the dimension of the phase space, both 
are hyperparameters. Hence, $l=T-(m-1)\tau$. With vectors $\mat{v}_{1}$, ..., $\mat{v}_{l}$, an RP image 
measures their pairwise distances, results in an $l\times l$ image whose element is
\begin{equation}\label{eq.rp}
\small
\begin{aligned}
\text{RP}_{i,j}=\Theta(\varepsilon - \|\mat{v}_{i} - \mat{v}_{j}\|),~~~1\le i,j\le l
\end{aligned}
\end{equation}
where $\Theta(\cdot)$ is the Heaviside step function, $\varepsilon$ is a threshold, and $\|\cdot\|$ is a norm function such as $\ell_{2}$ norm. Eq.~\eqref{eq.rp} 
generates a binary matrix with $\text{RP}_{i,j}=1$ if $\mat{v}_{i}$ and $\mat{v}_{j}$ are sufficiently similar, producing a black-white image ({\em e.g.}, Fig. \ref{fig.tsimage}(f)).

An advantage of RP is its flexibility in image size by tuning $m$ and $\tau$. Thus it has been used for time series classification 
\cite{silva2013time,hatami2018classification}, forecasting \cite{li2020forecasting}, anomaly detection \cite{lin2024hierarchical} and 
explanation \cite{kim2024cafo}. The method in \cite{hatami2018classification}, and similarly in HCR-AdaAD \cite{lin2024hierarchical}, omit the thresholding in Eq.~\eqref{eq.rp} and uses $\|\mat{v}_{i} - \mat{v}_{j}\|$ to produce continuously valued images 
to avoid information loss.

\subsection{Other Methods}\label{sec.othermethod}

Additionally, \cite{wang2015encoding} introduces Markov Transition Field (MTF). MTF is a matrix $\mat{M}\in\mathbb{R}^{Q\times Q}$ encoding transition probabilities among $Q$ time segments, to visualize UTS. ImagenTime \cite{naiman2024utilizing} stacks the delay embeddings $\mat{v}_{1}$, ..., $\mat{v}_{l}$ in Eq.~\eqref{eq.de} to an $l\times m\tau$ matrix for visualizing UTS. MSCRED \cite{zhang2019deep} applies heatmaps to $d\times d$ correlation matrices of $d$-variate MTS for anomaly detection. Earlier, \cite{kumar2005time} proposes Bitmaps to image discretized UTS and defines distances among them. Some methods use a mixture of imaging methods for richer representations, {\em e.g.}, \cite{wang2015imaging} stacks GASF, GADF, and MTF into a 3-channel image. FIRTS \cite{costa2024fusion} combines GASF, MTF, and RP. These multi-view representations have shown greater robustness than single-view images in these works for classification tasks.

\subsection{How to Model MTS}\label{sec.modelmts}

In the above methods, Heatmap ($\S$\ref{sec.heatmap}) can be 
used to visualize the 
variate-time matrices, $\mat{X}$, of MTS ({\em e.g.}, Fig. \ref{fig.structure}(b)), where correlated variates 
should be spatially close to each other. Line Plot ($\S$\ref{sec.lineplot}) can be used to visualize MTS by plotting all variates in the same image \cite{wimmer2023leveraging,daswani2024plots} or combining all univariate images to compose a bigger 
image \cite {li2023time}, but these methods only work for a small number of variates. Spectrogram ($\S$\ref{sec.spectrogram}), GAF ($\S$\ref{sec.gaf}), and RP ($\S$\ref{sec.rp}) were designed specifically for UTS. For these methods and Line Plot, which are not straightforward 
in imaging MTS, the general approaches 
include using channel independence assumption to model each variate individually \cite{nie2023time}, 
or stacking the images of $d$ variates to form a $d$-channel image 
\cite{naiman2024utilizing,kim2024cafo}. 
However, the latter does not fit some vision models pre-trained on RGB images which requires 3-channel inputs (more discussions are deferred to $\S$\ref{sec.processing}).

\vspace{0.2cm}

\noindent{\bf Remark.} As a summary, Table \ref{tab.tsimage} recaps the salient advantages and limitations of the five primary imaging methods that are introduced in this section.

\section{Imaged Time Series Modeling}\label{sec.model}

With image representations, time series analysis can be readily performed with vision models. This section discusses such solutions from 
traditional models to the SOTA models.

\begin{figure*}[!t]
\centering
\includegraphics[width=0.885\textwidth]{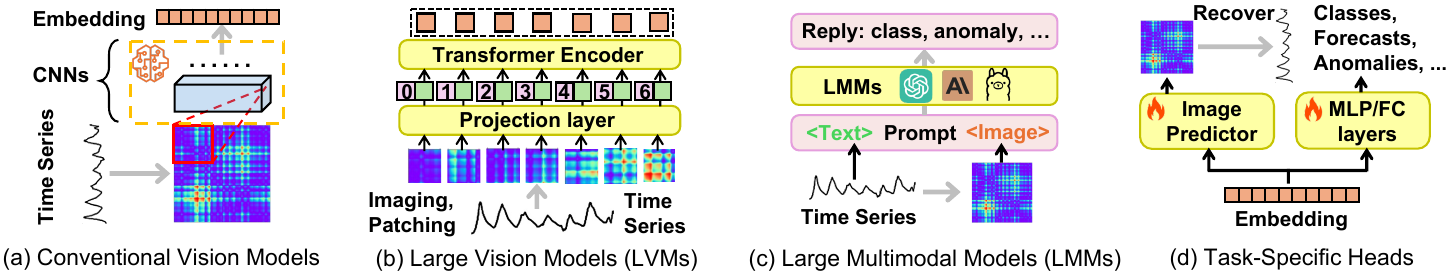}
\caption{An illustration of different modeling strategies on imaged time series in (a)(b)(c) and task-specific heads in (d).}\label{fig.models}
\end{figure*}

\subsection{Conventional Vision Models}\label{sec.cnns}

Following traditional 
image classification, \cite{silva2013time} applies a K-NN classifier on the RPs of time series, \cite{cohen2020trading} applies an ensemble of fundamental classifiers such as 
SVM and AdaBoost on the Line Plots 
for time series classification. As an image encoder, 
CNNs have been 
widely used for learning image representations. 
Different from using 1D CNNs on sequences 
\cite{bai2018empirical}, 
2D and 3D CNNs can be applied on imaged time series as shown in Fig. \ref{fig.models}(a). 
For example, 
regular CNNs have been used on Spectrograms \cite{du2020image} and Heatmaps \cite{wang2025timemixer}, tiled CNNs have been used on GAF images \cite{wang2015encoding,wang2015imaging}, dilated CNNs have been used on Heatmaps 
\cite{yazdanbakhsh2019multivariate}. More frequently, ResNet \cite{he2016deep}, Inception-v1 \cite{szegedy2015going}, and VGG-Net \cite{simonyan2014very} have been used on Line Plots \cite{semenoglou2023image}, Heatmap images \cite{zeng2021deep,wu2023timesnet}, RP images \cite{li2020forecasting,kim2024cafo}, GAF images \cite{barra2020deep,kaewrakmuk2024multi}, 
and even a mixture of GAF, MTF and RP images \cite{costa2024fusion}. In particular, for time series generation tasks, 
GAN frameworks of CNNs \cite{li2022tts,karami2024timehr} and a diffusion model with U-Nets \cite{naiman2024utilizing} have also been explored.

Due to their small to medium sizes, these models are often trained from scratch using task-specific training data. 
Meanwhile, fine-tuning {\em pre-trained vision models}  
have already been found promising in cross-modal knowledge transfer for time series anomaly detection \cite{namura2024training} and forecasting \cite{li2020forecasting}.

\subsection{Large Vision Models (LVMs)}\label{sec.lvms}

Vision Transformer (ViT) \cite{dosovitskiy2021image} has 
inspired the development of 
modern LVMs 
such as 
Swin \cite{liu2021swin}, BEiT \cite{bao2022beit}, and MAE \cite{he2022masked}. 
As Fig. \ref{fig.models}(b) shows, ViT splits an 
image into {\em patches} of fixed size, then embeds each patch and augments it with a positional embedding. The 
vectors of patches are processed by a Transformer 
as if they were token embeddings. Compared to CNNs, ViTs are less data-efficient, but have higher capacity. 
Thus, 
{\em pre-trained} ViTs have been explored for modeling 
imaged time series. For example, AST \cite{gong2021ast} fine-tunes DeiT \cite{touvron2021training} on the filterbank spetrogram of audios 
for classification tasks and finds 
ImageNet-pretrained DeiT is remarkably effective in knowledge transfer. The fine-tuning paradigm has also been 
adopted in \cite{zeng2023pixels,li2023time} but with different pre-trained models 
such as Swin by \cite{li2023time}. 
VisionTS \cite{chen2024visionts} 
attributes 
LVMs' superiority over LLMs in knowledge transfer 
to the small gap between the pre-trained images and imaged time series. 
It 
finds that with one-epoch fine-tuning, MAE becomes the SOTA time series forecasters on 
some benchmark datasets.

Similar to 
time series foundation models 
such as TimesFM \cite{das2024decoder}, 
there are some initial efforts in pre-training ViT architectures with imaged time series. Following AST, SSAST \cite{gong2022ssast} introduced a 
masked spectrogram patch prediction framework for pre-training ViT on a large dataset -- AudioSet-2M. Then it becomes a backbone of some follow-up works such as AST-SED \cite{li2023ast} for sound event detection. 
For UTS, ViTime \cite{yang2024vitime} generates a large set of Line Plots of synthetic UTS for pre-training ViT, which was found superior over TimesFM in zero-shot forecasting tasks on benchmark datasets.

\subsection{Large Multimodal Models (LMMs)}\label{sec.lmms}

As LMMs 
get growing attentions, some 
notable LMMs, such as LLaVA \cite{liu2023visual}, 
Gemini \cite{team2023gemini}, GPT-4o \cite{achiam2023gpt} and Claude-3 \cite{anthropic2024claude}, have been explored to consolidate the power of LLMs 
and LVMs 
for time series analysis. 
Since LMMs support multimodal input via prompts, methods in this thread typically prompt LMMs with 
textual and imaged representations of time series, 
and instructions on what tasks to perform ({\em e.g.}, Fig. \ref{fig.models}(c)).

InsightMiner \cite{zhang2023insight} is a pioneer work that uses the LLaVA architecture to generate 
texts describing the trend of each input UTS. It extracts the trend of a UTS by Seasonal-Trend decomposition, encodes the Line Plot of the trend, and concatenates the embedding of the Line Plot with the embeddings of a textual instruction, which includes a sequence of numbers representing the UTS, {\em e.g.}, ``[1.1, 1.7, ..., 0.3]''. The concatenated embeddings are taken by a language model for generating trend descriptions. 
Similarly, \cite{prithyani2024feasibility} adopts the LLaVA architecture, but for MTS classification. An MTS is encoded by 
the visual 
embeddings of the stacked Line Plots of all variates. 
The matrix of the MTS is also verbalized in a prompt 
as the textual modality. 
By integrating token embeddings, both 
methods fine-tune some layers of the LMMs with some synthetic data.

Moreover, zero-shot and in-context learning performance of several commercial LMMs have been evaluated for audio classification \cite{dixit2024vision}, anomaly detection \cite{zhuang2024see}, and some synthetic tasks \cite{daswani2024plots}, where the image 
and textual representations of a query 
time series are integrated into a prompt. For in-context learning, these methods inject the images of a few example time series and their labels ({\em e.g.}, classes) 
into an instruction to prompt LMMs for assisting the prediction of the query time series.

\subsection{Task-Specific Heads}\label{sec.task}

As Fig. \ref{fig.models}(d) illustrates, for classification tasks, most of the methods in Table \ref{tab.taxonomy} adopt a fully connected (FC) layer or multilayer perceptron (MLP) to transform an embedding into a probability distribution over all classes. For forecasting tasks, there are two approaches: (1) using a $d_{e}\times W$ MLP/FC layer to directly predict (from the $d_{e}$-dimensional embedding) the time series values in a future time window of size $W$ \cite{li2020forecasting,semenoglou2023image}; (2) predicting the pixel values that represent the future part of the time series and then recovering the time series from the predicted image \cite{yang2023your,chen2024visionts,yang2024vitime} ($\S$\ref{sec.processing} discusses the recovery methods). Imputation and generation tasks resemble forecasting 
as they also predict time series values. Thus approach (2) has been used for imputation \cite{wang2015imaging} and generation \cite{naiman2024utilizing,karami2024timehr}. 
When using LMMs for classification, text generation, and anomaly detection, most of the methods prompt LMMs to produce the desired outputs in textual answers, circumventing task-specific heads \cite{zhang2023insight,dixit2024vision,zhuang2024see}.


\section{Pre-Processing and Post-Processing}\label{sec.processing}

To be successful in using vision models, some subtle design desiderata 
include {\bf time series normalization}, {\bf image alignment} and {\bf time series recovery}.

\vspace{0.1cm}

\noindent{\bf Time Series Normalization.} Vision models are usually trained on 
standardized images. To be aligned, the images introduced in $\S$\ref{sec.tsimage} should be normalized with a controlled mean and standard deviation, as did by \cite{gong2021ast} on spectrograms. In particular, as Heatmap is built on raw time series values, the commonly used Instance Normalization (IN) \cite{kim2022reversible} can be applied on the time series as suggested by VisionTS \cite{chen2024visionts} since IN share similar merits as Standardization. 
Using Line Plot requires a proper range of y-axis. In addition to rescaling time series 
\cite{zhuang2024see}, ViTST \cite{li2023time} introduced several methods to remove extreme values from the plot. GAF requires min-max normalization on its input, as it transforms time series values withtin $[0, 1]$ to polar coordinates ({\em i.e.}, arccos). In contrast, input to RP is usually normalization-free as an $\ell_{2}$ norm is involved in Eq.~\eqref{eq.rp} before thresholding.

\vspace{0.1cm}

\noindent{\bf Image Alignment.} When using pre-trained models, it is imperative to fit the image size to the input requirement of the models. This is especially true for Transformer based models as they use a fixed number of positional embeddings to encode the spatial information of image patches. For 3-channel RGB images such as Line Plot, it is straightforward to meet a pre-defined size by adjusting the resolution when producing the image. For images built upon matrices such as Heatmap, Spectrogram, GAF, RP, the number of channels and matrix size need adjustment. For the channels, one method is to duplicate a matrix to 3 channels \cite{chen2024visionts}, another way is to average the weights of the 3-channel patch embedding layer into a 1-channel layer \cite{gong2021ast}. For the image size, bilinear interpolation is a common method to resize input images \cite{chen2024visionts}. Alternatively, AST \cite{gong2021ast} 
resizes the positional embeddings instead of the images to fit the model to a desired input size. However, the interpolation in these methods may either alter the time series or the spatial information in positional embeddings.

\vspace{0.1cm}

\noindent{\bf Time Series Recovery.} As stated in $\S$\ref{sec.task}, tasks such as forecasting, imputation and generation requires predicting time series values. For models that predict pixel values of images, post-processing involves recovering time series from the predicted images. Recovery from Line Plots is tricky, it requires locating pixels that 
represent time series and mapping them back to the original values. This can be done by manipulating a grid-like Line Plot as introduced in \cite{yang2023your,yang2024vitime}, which has a recovery function. In contrast, recovery from Heatmap is straightforward as it directly stores the predicted time series values \cite{zeng2021deep,chen2024visionts}. Spectrogram is underexplored in these tasks and it remains open on how to recover time series from it. The existing work \cite{zeng2023pixels} uses Spectrogram for forecasting only with an MLP head that directly predicts time series. 
GAF supports accurate recovery by an inverse mapping from polar coordinates to normalized time series \cite{wang2015imaging}. However, RP lost time series information during thresholding (Eq.~\ref{eq.rp}), thus may not fit recovery-demanded tasks without using an {\em ad-hoc} prediction head.

\section{Challenges and Future Directions}\label{sec.future}

\noindent{\bf Fundamental Understanding.} With various imaging methods available, most existing works select them based on intuition. There remains a gap in both theoretical and empirical understanding of questions such as which imaging methods fit which tasks and whether LVMs truly learn patterns that make them better suited to time series than LLMs. Some existing works evaluate multiple imaging methods, but in limited tasks. For example, ImagenTime \cite{naiman2024utilizing} compares the representation abilities of GAF, STFT, and delay embedding ($\S$\ref{sec.othermethod}) in a time series generation task. However, a thorough understanding to guide the development of LVMs and LMMs across imaging methods is absent. This survey provides an initial comparative discussion in $\S$\ref{sec.tsimage}. Further empirical and theoretical investigations are essential to the synergy between LVMs/LMMs and time series analysis.

\vspace{0.1cm}

\noindent{\bf Modeling the Correlation of Variates in MTS.} $\S$\ref{sec.modelmts} reviewed existing methods for imaging MTS, each with limitations. For example, Heatmaps ({\em e.g.}, Fig. \ref{fig.tsimage}(b)) encode spatial relationships, so the row order of variates affects how correlations are modeled, implying correlated variates should be spatially close to each other. Similarly, Line Plots do not explicitly capture inter-variate correlations. Stacking one channel per variate into a $d$-channel input hinders the use of pre-trained LVMs, which expect 3-channel RGB inputs. Therefore, there is a need for more effective imaging or modeling techniques ({\em e.g.}, incorporating graph neural networks (GNNs) on variates) to better capture correlations in MTS.

\vspace{0.1cm}

\noindent{\bf Advanced Imaging for Time Series.} In addition to the basic methods introduced in $\S$\ref{sec.tsimage}, advanced image representations hold promise. For example, InsightMiner \cite{zhang2023insight} adopts seasonal-trend decomposition, which is often used to extract components that can serve as inductive biases for time series models. Extending this idea to decompose images such as Spectrogram, GAF, and RP into finer-grained components may enhance vision models' effectiveness. Moreover, combining multiple imaging methods from different views, such as frequency (Spectrogram), temporal structure (GAF), and recurrence patterns (RP), can provide richer representations. FIRTS \cite{costa2024fusion} stacks images across channels for a classification task, but is limited to images of the same size. Modeling a mixture of arbitrary images by methods such as multi-view learning may enable more flexibility.

\vspace{0.1cm}

\noindent{\bf Multimodal Time Series Models and Agents.} As can be seen from Table \ref{tab.taxonomy}, the existing research on multimodal analysis (with vision modality) is much less than unimodal analysis, with a limited scope of time series tasks. Given the existing LLMs for time series such as Time-LLM \cite{jin2024time} and S\textsuperscript{2}IP \cite{pan2024s}, it is appealing to introduce vision modality 
to further boost the performance in wide tasks such as forecasting, classification and anomaly detection. Furthermore, the visual representation of time series provides the foundation for exploring multimodal AI agents \cite{xie2024large} for more intricate and nuanced tasks that requires reasoning and interactions with environments, such as 
root cause analysis in AI for IT Operations (AIOps).

\vspace{0.1cm}

\noindent{\bf Vision-based Time Series Foundation Models.} A foundation model (FM) is a deep learning model trained on vast datasets that is applicable to a wide range of tasks. Recent time series FMs, such as TimesFM \cite{das2024decoder}, MOMENT \cite{goswami2024moment}, Chronos \cite{ansari2024chronos} and Time-MoE \cite{shi2024time}, are mostly built upon LLM architectures and trained on raw time series. Given the potential of image representation, 
it is promising to explore vision models as a new architecture to revolutionize time series FMs. This research direction not only leverages the advantages of LVMs as introduced in $\S$\ref{sec.introduction} ({\em e.g.}, the 
prior knowledge extracted from the vast pre-training images), but also enables future development of vision-language FMs for time series.

\section{Conclusion}

In this paper, we present the first survey on leveraging vision models for time series analysis, whose general process structures the survey. We propose a new taxonomy consisting of imaging and modeling methods for time series. We discuss the pre- and post-processing steps as well. Each category encompasses representative methods and relevant remarks. The survey also highlights the challenges and future directions for further advancing time series analysis with vision models.

\clearpage

\section*{Acknowledgments}

This work was partially supported by NSF (2134079), CAREER Award IIS-2338878, IIS-2331908, and research gifts from NEC Laboratories America and Morgan Stanley.

\small
\bibliographystyle{named}
\bibliography{ref}
\normalsize

\end{document}